\documentclass{article}

\usepackage{microtype}
\usepackage{graphicx}
\usepackage[caption=false,font=footnotesize]{subfig}
\usepackage{booktabs} 

\usepackage[T1]{fontenc}
\usepackage{multirow}

\usepackage{hyperref}

\usepackage{import}
\usepackage{tikz}
\usetikzlibrary{positioning}

\usepackage[accepted]{icml2018}

\icmltitlerunning{Blind Pre-Processing: A Robust Defense Method Against Adversarial Examples}

\begin{document}

\twocolumn[
\icmltitle{Blind Pre-Processing: A Robust Defense Method Against Adversarial Examples}



\icmlsetsymbol{equal}{*}

\begin{icmlauthorlist}
\icmlauthor{Adnan Siraj Rakin}{to}
\icmlauthor{ Zhezhi He}{to}
\icmlauthor{Boqing Gong}{goo}
\icmlauthor{Deliang Fan}{to}
\end{icmlauthorlist}

\icmlaffiliation{to}{Department of Computer Engineering, University of Central Flordia, Orlando, USA}
\icmlaffiliation{goo}{ Tencent AI Lab at Seattle, Bellevue WA 98004}

\icmlcorrespondingauthor{Deliang Fan}{dfan@knights.ucf.edu}

\icmlkeywords{Machine Learning, ICML}

\vskip 0.3in
]



\printAffiliationsAndNotice{}  

\begin{abstract}
Deep learning algorithms and networks are vulnerable to perturbed inputs which is known as adversarial attack. Many defense methodologies have been investigated to defend against such adversarial attack. In this work, we propose a novel methodology to defend the existing powerful attack model. We for the first time introduce a new attacking scheme for the attacker and set a practical constraint for white box attack. Under this proposed attacking scheme we present the best defense ever reported against some of the recent strong attacks. It consists of a set of non linear function to process the input data which will make it more robust over adversarial attack. However, we make this processing layer completely hidden from the attacker. Blind pre-processing improves the white box attack accuracy of MNIST from 94.3\% to 98.7\%. Even with increasing defense when others defenses completely fails, blind pre-processing remains one of the strongest ever reported. Another strength of our defense is that, it eliminates the need for adversarial training as it can significantly increase the MNIST accuracy without adversarial training as well. Additionally, blind pre-processing can also increase the inference accuracy in the face of powerful attack on Cifar-10 and SVHN data set as well without much sacrificing clean data accuracy.
\end{abstract}

\section{Introduction}
Deep Neural Network (DNN) has achieved great success in performing classification \cite{andor2016globally}, image detection \cite{he2015delving} and speech recognition \cite{xiong2016achieving}. All these successes lead to another interesting topic about how well DNN will perform practically depends heavily on how well we address the issues concerning the security of these complex models. It was observed by  Szegedy et al. (2014) for computer vision \cite{goodfellow2014explaining} and Biggio et al. (2013) for malware detection \cite{biggio2013evasion} that DNNs are vulnerable to adversarial examples which can be generated just by slightly changing inputs or by introducing random noise to the inputs. Not only image classification but also other popular fields of DNN are facing vulnerability to adversarial examples \cite{kos2017adversarial,kos2017delving}. Similarly, The behaviour of popular Convolutional Neural Network (CNN) models is no exception as it shows no resistance to adversarial examples. 

\begin{figure}[h]
\centering
\scalebox{1}{
\includegraphics[width=8cm,height=3cm]{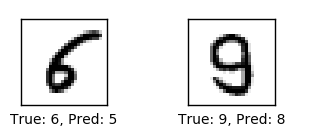}
}
\caption{Adversarial attack causing miss classification in MNIST data set}
\end{figure}

There are now various works on generating adversarial attacks and developing corresponding defense methods. Attacking CNN can be classified into two major categories. In one case the attacker has full access to the network model and parameter which is known as white box attack. On the other hand, for black box attack the attacker sees the model as a black box and can only play with the data. Fast Gradient sign method is one of the popular white box attack method which uses the sign of the gradient of loss function \cite{goodfellow2014explaining}. Different authors have proposed different attack models to adapt the attacking techniques beating the improving defenses \cite{szegedy2013intriguing,kurakin2016adversarial,kos2017delving,papernot2016limitations,moosavi2016deepfool}. Some of the recent attack methods achieved even better success causing CNN to classify whole MNIST data set to wrong classes. For example, projected gradient descent is one of the most powerful white box attack till date \cite{mkadry2017towards}. 

Many defense method have been proposed to beat adversarial attacks. Some of them achieved good successes mainly against weak attacks, such a Fast Gradient Sign Method(FGSM) and Jacobian Slaiency Map attack (JSMA) \cite{gu2014towards,papernot2016distillation,papernot2016towards,xu2017feature}. However, None of them could recover current state of the art accuracy against powerful attacks. One major improvement to the those defenses came when defending model started using adversarial examples while training \cite{tramer2017ensemble,mkadry2017towards}. But this defense has to two major drawbacks. First, as a defender we do not know the attack model. So it is difficult to choose which adversarial attack method should be used to generate adversary for training. Second, it would double training cost. Still most recent defenses use this technique to strengthen their defense. In this work we look to eliminate the need for using adversarial training by processing the input with the objective to make the neural network more robust. Moreover, we observed all these defenses mentioned here loses its strength when the attack strength is increased. So by implementing this proposed defense we look to make the defense more robust against a wide variety of attacks.

Inspired by recent One hot coding and thermometer encoding of input data \cite{buckman2018thermometer}, in this work, we propose blind pre-processing (BP) of input data. We process the input data using Tanh function, batch normalization, thermometer encoding and one hot encoding. These techniques work in combination with discrete data to defend CNN against adversarial attacks. However, we change the attack set up by introducing an engineering solution to the gradient problem suggested by recent Obfuscated Gradients paper \cite{obfuscated-gradients}. From the work of obfuscating gradient it is obvious that no white box defense model would work in the presence of powerful attack model. Specifically, none of them would work if it has gradient shattering, stochastic gradient and vanishing gradient as their defense weapon. They also suggested thermometer encoding is using stochastic gradient and they could fool this defense by their proposed attacking model. However, we introduce a more practical white box attack where the attacker will have access to the model, weights, input, gradient and training algorithm but not to the pre-processing layer. we will use obfuscated logic design by proposing a hardware assisted pre-processing layer to counter the obfuscated gradient's false sense of security \cite{obfuscated-gradients}. We strongly believe their attack model has too many freedom and restricting the defense with a lot of constraints. As a result, we believe future attack model should evaluate their models strength by considering this blind processing layer in front of the model. This defense is more realistic for practical application such as autonomous vehicles.

Our proposed defense method makes the CNN more robust to defend the strongest attack, Logit Space Gradient Ascent(LS-PGA) attack and also the general strong attack suggested by Anish et. al \cite{obfuscated-gradients}. It is a modified version of PGD attack on discrete data. The main contribution of this work is that we could achieve very high accuracy on MNIST without adversary training for the first time. If we include adversary training, our defense methodology provides the best accuracy against MNIST, Cifar 10 and SVHN till date. Not only that with increasing attack strength when all the defenses completely fails, blind pre-processing still defends the CNN causing minimal accuracy loss.  Contributions of this work are summarized as follows:

\begin{itemize}
    \item We derived from the previous analysis of adversarial examples that introducing non linearity would increase the defense to any adversarial attack. As a result, We introduced a series of functions performing non linear transformation of input data to build a robust defense.
    \item After making the model robust using pre-processing we propose obfuscated logic based implementation of the pre-processing layer so that this network do not fail due to its obfuscated gradients. This modification will also make all the defense that use obfuscated gradient to succeed again.
    \item This proposed defense resulted in increasing the accuracy for MNIST data set from 94.02\% to ~98.66\% with adversarial training and from 0\% to 98.29\% without adversarial training. This is the best so far achieved on LS-PGA attack without adversarial examples. Even with adversarial training, previous attempts could only reach as far as ~95\%. 
    \item We showed that with increasing attack strength our defense remains one of the strongest. Our analysis provides a more detailed understanding of why blind pre-processing stands out in the advent of adversary attack while others fail.
    \item Additionally by using Adversary training we could achieve 95.06\% accuracy on SVHN data set and 86.66\% accuracy on Cifar 10 data set.
    
\end{itemize}

In the next two sections we summarize the attack model and blind processing scheme to defend those attack models. In section IV our proposed defense technique is described and the following two sections represents the results and analysis of the proposed BP method.

\section{Attack Algorithm}

In this work, we assume that the adversary has complete access to the network which is popularly known as the white box attack. It was observed from previous works that if a defense model performs well for white box attack it should naturally perform better against black box attack \cite{mkadry2017towards}. Since black box attack becomes really weak due to the unavailability of network parameters. That is why in general any defense that wants to prove its effectiveness must perform well against white box attack. In order to formulate the attack method for this work, we first investigate some of the popular white box attack models:

Madry et al. introduced Projected gradient descent which achieved 100\% success in fooling CNN to miss classify MNIST dataset  \cite{mkadry2017towards}. In their work they show that PGD will generate universal adversary among the first order approaches. They suggested that training network with this kind of adversary would make the neural network more robust. Taking their work forward this algorithm was modified for discrete input \cite{buckman2018thermometer} known as discrete gradient ascent and logit spcae projected gradient ascent.

\begin{algorithm}
   \caption{LS-PGA Attack}
    \begin{algorithmic}
        \STATE Input: Image(x),Label(l), attack steps(n), $\epsilon,\delta$,k(quantized level),Loss function L(w,f(x),l)
        \STATE mask = 0
        \STATE $low =max[0,x-\epsilon]$
        \STATE $high=min[0,x+\epsilon]$
        \FOR{$i = 1$ to ${k+1}$}
            \STATE $mask=mask+f_{sdi}(\alpha*low+(1-\alpha*high)$
        \ENDFOR
        \STATE $u^0\ Initialization\ using\ mask$ 
        \STATE $T=1$ 
        \STATE $z^0=F(\sigma(u^0/T))$ 
        \FOR{$j = 1$ to ${n}$}
            \STATE $grad=\nabla_{z^{j-1}}.L(w,z^{j-1},y)$
            \STATE $u^j=u^{j-1}+\epsilon.grad$
            \STATE $z^{j}=\sigma(z^{j-1}/T)$
            \STATE $T=T.\delta$
        \ENDFOR
        \STATE {\bfseries output:}\ z \ after\ n\ iteration 
    \end{algorithmic}
\end{algorithm}

Since this method is more  suitable for discrete inputs we choose to evaluate our model against this attack. Additionally the success of this white box attack tempted us to evaluate our defense model against it. Finally, based on the success of these attacks against varying defenses we summarize their strength in table 1.

\begin{table}[]
\centering
\caption{Classification of the strength of the attacks}
\scalebox{0.85}{
\begin{tabular}{|l|l|l|l|l|}
\hline
Attack                                                                      & FGSM & JSMA & C \& W                                                                           & PGD                                                                                                                                    \\ \hline
\begin{tabular}[c]{@{}l@{}}List of \\ defense\\ it can \\ beat\end{tabular} & x    & x    & \begin{tabular}[c]{@{}l@{}}1.Distillation\\ 2. Feature \\ Squeezing\end{tabular} & \begin{tabular}[c]{@{}l@{}}1.Distillation\\ 2. Feature \\ squeezing\\ 3. PGD \\ training\\ 4.Thermometer \\ (Without adv)\end{tabular} \\ \hline
Strength                                                                    & Weak & Weak & Medium                                                                           & Strong                                                                                                                                 \\ \hline
\end{tabular}}
\end{table}

The vast majority of machine learning problem is solved with first order  methods like gradient descent or other closely modified versions. That is why those attacks that rely on only first order information could be labeled as universal attack. Since LS-PGA/PGD attack model depends only on the first order information, defending against LS-PGA attack would certainly make the model robust against a wide variety of attack models. This attack takes place by placing each pixel in a random bucket within $\epsilon$.
At each step of the attack it will look at that bucket to find values within $\epsilon$ of the true value and select value that will do the most harm. The outcome of this attack will vary depending on the initialization. So the attack needs to be run several times to get the desired result. In case of Logit space projected gradient ascent first discrete encoding was relaxed to continuous space to perform projected gradient ascent. By changing the values of $\epsilon$  and attack step we can certainly change the strength of the attack. We report the performance of our defense against varying attack strength in section VI. 

Recent attack developed by \cite{obfuscated-gradients} is reported to be the more generalized and powerful till now. They not only fool the thermometer encoding but also all the recent defenses. In the next section based on their analysis we introduce the concept of blind processing and also present why obfuscated gradient can actually work as a defense under our defense model. Moreover, we present a attacking scheme for future attacking models to test and evaluate their attack models.

\section{Concept of Blind Pre-Processing}
Obfuscated gradient occur due to gradient shattering, stochastic gradient and gradient vanishing \cite{obfuscated-gradients}. From the analysis presented by \cite{obfuscated-gradients}, it is evident any defense that uses one of these as their weapon to defend the adversary that will eventually fail. As reported this obfuscated gradient gives a false sense of security. Their generalized attack have caused all of the recent defense models to fail. Thermometer encoding is no exception as it has both gradient shattering and stochastic gradient issue. For our defense to work we need to make this obfuscated gradient based defense to function properly in the advent of powerful attack. So the final conclusion is in a typical white box attack set up it is impossible to defend white box adversarial attack. Even though \cite{mkadry2017towards} defense works but the success is very small. We conclude that the attacker has too many freedom and defense is constrained by different optimization problems. As a result, we propose a more practical way of dealing with the problem on hand. We propose blind processing where this processing layer is absolutely unknown to the attacker. 

In this defense model, the attacker can not access the function of the pre-processing layer. We use obfuscated logic to hide the design of our pre-processing layer. The attacker can access the input and output of the process but can never predict the functionality of the model. Previous work have been done in this field using hybrid polymorphic logic gate to obscure the functionality of a logic gate. We propose using such logic gate to pre-process the data in a practical application such as autonomous vehicle. In this way the attacker can never replicate the module for two reason. Firstly, the layout using polymorphic logic looks same for all kinds of logic. In this way the hacker can never predict the functionality by hacking the module and analyzing layout. Secondly, the attacker can never guess the functionality using the pattern of input and output of the pre-processing layer because this logic is key assisted. In this way whoever does not have the key can never access the pre-processing module to generate input output data. The user have always the power to use this particular key to train the model and can switch off this particular functionality using another key. Additionally, the user can also use different key for different pre-processing functions. As a result it would become impossible for the attacker to guess the functionality of the defense model in the inference mode as well. In this way the whole pre-processing layer becomes a blind pre-processing layer for the attacker. Additionally, the key of this kind of logic is also secured. If the pre-processing layer has an input of M and logic gate of N then total number of key would be $2^{M+3N}$. So for a typical MNIST classifier the input to that function would be 256 and even if we assume the pre-processing layer is a simple and operation it still would have close to $9X10^{77}$ possible key combination. practically this is impossible for the attacker to hack in a limited amount of time. From this proposal we assure the security of the pre-processing layer which we define as blind pre-processing. As suggested by \cite{obfuscated-gradients} that a defence must define the attackers accessibility to make the defenses more practical. In our case, blind pre-processing keeps all the other attribute of the white box attack in tact. The attacker still has access to the following information:
\begin{enumerate}
    \item Model architecture 
    \item weights and gradients
    \item Training algorithm 
    \item Training data before and after pre-processing
\end{enumerate}

However, we are denying two accessibility to the attacker. Firstly, the attacker can not predict the pre-processing layer nor access the layer for gradients. Secondly access query is not allowed without the key in the pre-processing layer. This two attributes constitute the blind pre-processing.

\begin{figure}[h]
\centering
\scalebox{1}{
\includegraphics[width=8cm]{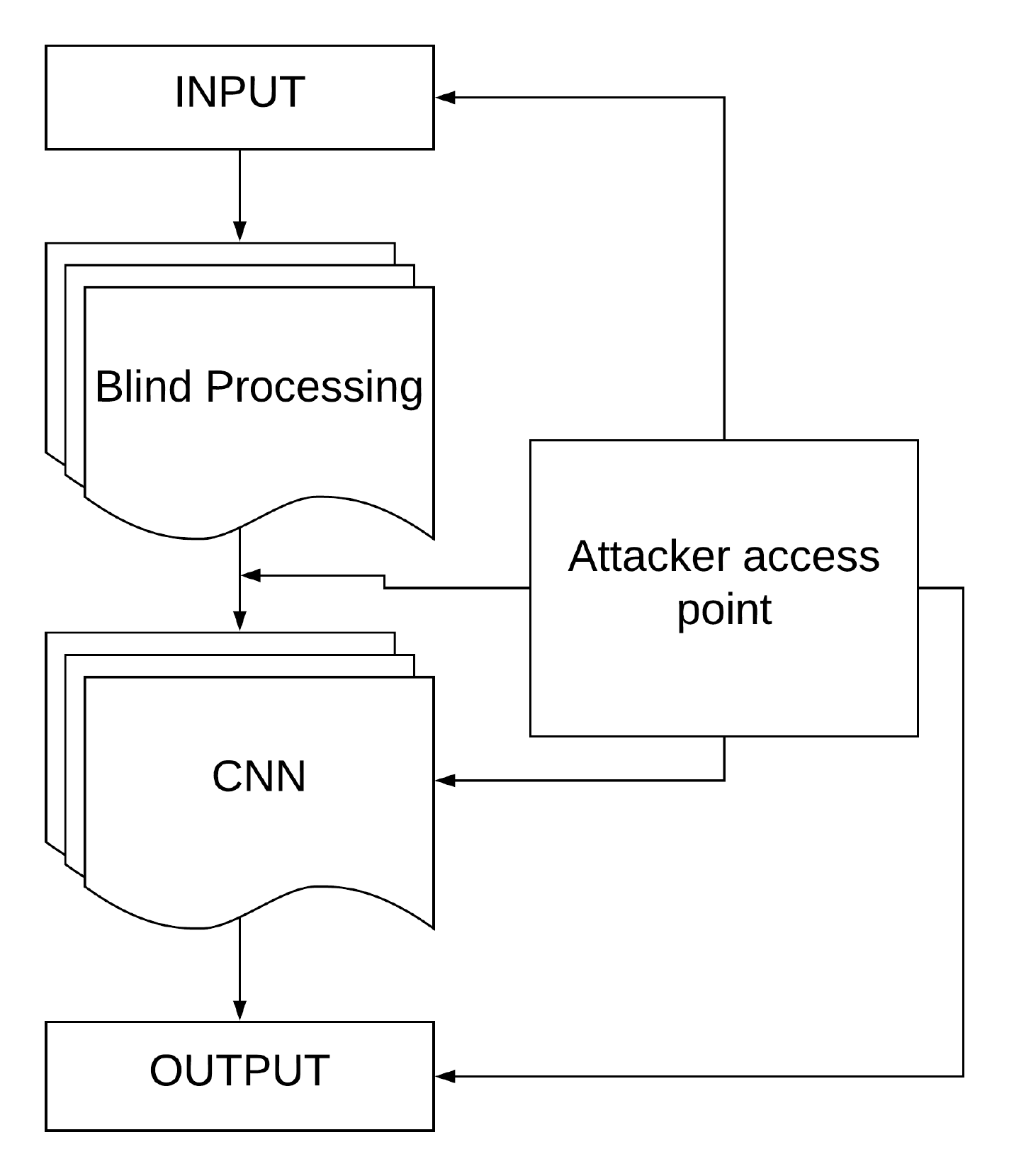}
}
\caption{Attacker access zone under Blind Pre-Processing defense}
\end{figure}

From the analysis of \cite{obfuscated-gradients} we concluded that it is impossible for traditional defenses to defend under the previous attacking scheme. As a consequence we introduce the concept of blind pre-processing to encounter the obfuscated gradient issue. As a result we set a new attacking scheme to evaluate their attack model against our defense model. We strongly claim that with blind pre-processing the generalized attack method suggested by \cite{obfuscated-gradients} becomes a failure. We suggest future attack model to generate adversary without back propagating through the pre-processing layer or generate adversary just by using the data after pre-processing layer. Since this new scheme for white box attack is more practical and gives the attacker some form of challenge. Thus any attack that want to succeed must prove its worth by beating these obfuscated gradient based defenses without approximating their gradient. Under this blind pre-processing defense model approximating the pre-processing function is close to impossible let alone the gradient.

\section{Proposed Defense Method}
In the analysis of existence of adversary in deep neural network, goodfellow et al. \cite{goodfellow2014explaining} concluded that deep neural networks exhibit vulnerability to adversarial examples due to their extreme linearity. Linearity of these models is the reason why they can not resist adversary. For example MNIST data set has an input precision of 8 bit. As a result, a common notion would be why a network would respond differently when the input is x+ $\epsilon$ instead of x where $\epsilon$ is really small. However, as practical experiments show that they do behave differently to these two inputs. To understand this we have to look into the linearity theory of deep models. Output of neural network y could be represented as y= $\sigma(W^TX')$.

\begin{equation}
{W^TX'=W^TX+W^T\epsilon}
\end{equation}

Even if $\epsilon$ is really small, if a weight matrix has n dimension and an average value of m, then this perturbation would result in $\epsilon$*m*n increase of activation. Thus with increased dimension, this noise keeps increasing linearly. It suggests that, with sufficient large input dimension, a network will always be vulnerable to adversary. It would then provide two possible directions to solve this issue. One is to introduce non-linearity to the network and the other is to eliminate unnecessary information from input data. First one is a bit tricky as introducing non-linearity inside the neural network creates problem for calculating gradient. Previously, Different non-linear activation functions were investigated but none of them worked due to the difficulty of computing gradients and poor accuracy on real test data set. So in this work, we choose the second direction by adding a pre-processing layer in front of neural network model which will not have any gradient issue. Additionally, blind processing makes the obfuscated gradient issue suggested by \cite{obfuscated-gradients} nullified. Moreover, adding pre-processing layer in front of the CNN model does not hamper the accuracy on clean data. 

Our proposed blind pre-processing (BP) has tanh and Batch Normalization (BN) to process the input data in an attempt to make the CNN more robust. As shown in fig. 2 The input data goes through a non linear transformation first which is tanh function. A batch normalization layer is also added which is similar to the BN layer in CNN but it does not have any learnable parameters. After blind pre-processing input data were quantized to 15 different levels. They were presented in vectorized form of one hot coded value. Finally, they were encoded using thermometer encoding which basically inverts all the digits after first one appears. We observe that adding blind pre-processing before qunatization made the defense model more robust. So during all our experiment we quantize the input data after blind pre-processing. In our experiment it was found that adversarial training was no longer necessary for MNIST.

We also believe those function that uses local information works well as pre-processing layer. Moreover, this pre-processing can not sacrifice much accuracy on the clean data at the same time. Previously, thermometer encoding and quantization were suggested as some form of pre-processing. Our proposed blind pre-processing layer consists of tanh and batch normalization layer. But there could be other possible function that can be used as a blind pre-processing layer. The description of the functionality of those layers are presented here:
\begin{enumerate}
    \item Tanh and Sigmoid Function: The most popular activation functions are Sigmoid and Tanh functions. Both functions are used in typical deep neural networks. Mathematically they can be expressed as:
    \begin{equation}
    F_{tanh}(x)=\frac{e^x-e^{-x}}{e^x+e^{-x}} 
    \end{equation}
    \begin{equation} F_{sigmoid}(x)=\frac{1}{1+e^{-x}} 
    \end{equation}

Again during experiments it was observed that tanh outperforms sigmoid function. So we choose tanh to do the non linear transformation in the first layer.
    \item Batch Normalization (BN) function: Our normalization layer is similar to the batch normalization layer in a typical Convolutional Neural Network. However, it does not have any learnable parameters. Rather, this is just a pre-processing layer in front of the original neural network model.The function can be written as this: 
    \begin{equation}
\small Y=\frac{X-\mu}{{\sigma}}
\end{equation}
where $Y$ and $X$ denote the input and output tensor respectively. $\sigma$, $\mu$ represents the standard deviation and mean of the input data for that particular batch. This is a linear layer which basically makes the mean of the input data to be zero and standard deviation to one.
\end{enumerate}

We used tanh in place of sigmoid in this work. As these set of combinations were found to perform better than the other. Each input image is passed through a tanh layer which works as a filter. Followed by a 3 by 3 filter of Maximum smoothing. Then the data goes through a Batch Normalization layer followed by a 15 level quantization. Finally encoded values were feed into the model for training.

\section{Experimental Results}
\subsection{Performance in MNIST Data set}

Our proposed defense method was first tested on a simple LeNet architecture for MNIST data set.MNISt is a set of hand written digit with 55000 training data and 10000 testing data. We used LS-PGA attack model to evaluate our model as described in section 2. In LS-PGA, we set $\epsilon$ =0.3, $\delta=1.2$, xi=1 and step-of-attack = 7 for the following experiments. Here, $\epsilon$ indicates how much the input pixels can be changes and $\delta$ is the attack step as described in section II.  In the later section, we also varied such parameters to see the effects on the performance of our proposed defense method.

Our defense model uses various processing layers to defend against adversary. We here investigate the efficacy of individual and combination of different pre-processing layers. Tanh function was chosen to be the first layer of  blind pre-processing. It was found that the performance is better if we put tanh function before batch normalization layer. 
\begin{table}[h]
\centering
\caption{Result of Sigmoid and Tanh Functions}
\label{coding}
\scalebox{0.8}{
\begin{tabular}{@{}cccc@{}}
\toprule
\textbf{Function}  & \textbf{Clean training} & \textbf{ADV training} \\ \midrule
\textbf{Sigmoid} & \textbf{0} & \textbf{8} \\ \midrule
\textbf{tanh} & \textbf{69.55} & \textbf{94.42} \\ 
\bottomrule

\end{tabular}}
\end{table}

Table 2 indicates that the performance of tanh was far better than sigmoid as a filtering function. As a result, for our experiment we choose tanh instead of sigmoid. Meanwhile, thermometer encoding fails completely when adversarial training was removed. Whereas tanh function improved the accuracy even without adversary training. After tanh, we add a smoothing layer to investigate the effect of this layer. Based on the experiment, we observe maximum smoothing performs better than average smoothing.

\begin{figure}[h]
\centering
\scalebox{1}{
\includegraphics[width=8cm]{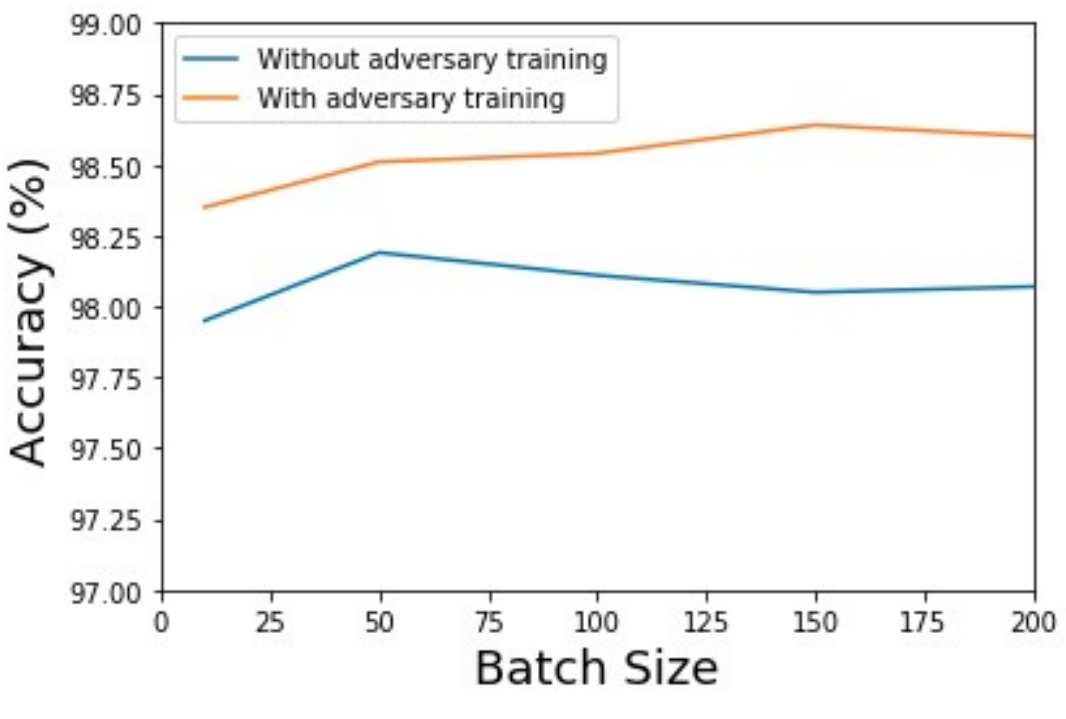}
}
\caption{Effect of changing the batch size both with and without adversarial training}
\end{figure}

 

One of the key function here is the batch normalization layer. As a result, the batch size becomes an important tuning parameter for this defense to optimize. We also investigated the defense efficacy using varying batch size on blind pre-processing. Increasing the batch size would increase the accuracy of the overall data set. Because if the batch size is one, batch normalization is the same as normalizing the whole image. In experiments, we found that there is an optimal batch size to achieve highest accuracy without adversary training as shown in fig. 3. When adversarial training is included, in general, the accuracy increases when increasing batch size. For the case of no adversary training, we found that an optimum batch size of 50, which leads to the best accuracy. However, looking at the clean training we decided to choose an optimum batch size of 100 for our experiments. Because in figure 4 the red line goes up a little bit while blue line goes down from 50 to 100 batch size. It means we could optimize the accuracy between with adversary training and without adversary training using a batch size of 100 in place of 50.

\begin{table}[]
\centering
\caption{Accuracy on MNIST dataset using different defense combination}
\label{my-label}
\begin{tabular}{|l|l|l|l|l|}
\hline
\begin{tabular}[c]{@{}l@{}}Train\\ Type\end{tabular}                                 & \begin{tabular}[c]{@{}l@{}}Work\\ Type\end{tabular}                      & Defense                                                          & Clean & Attack \\ \hline
\multicolumn{2}{|l|}{Original CNN}                                                                                                                              & No defense                                                       & 99.45 & 0      \\ \hline
\multirow{5}{*}{\begin{tabular}[c]{@{}l@{}}Without \\ Adv. \\ Training\end{tabular}} & \multirow{3}{*}{\begin{tabular}[c]{@{}l@{}}Proposed\\ BP\end{tabular}}   & BN                                                               & 99.23 & 98.05  \\ \cline{3-5} 
                                                                                     &                                                                          & Tanh                                                             & 99.56 & 69.55  \\ \cline{3-5} 
                                                                                     &                                                                          & Tanh \& BN                                                       & 99.2  & 98.3   \\ \cline{2-5} 
                                                                                     & \multirow{2}{*}{\begin{tabular}[c]{@{}l@{}}Previous\\ Work\end{tabular}} & \begin{tabular}[c]{@{}l@{}}Thermometer\\ Encoding\end{tabular}   & 99.2  & 0      \\ \cline{3-5} 
                                                                                     &                                                                          & \begin{tabular}[c]{@{}l@{}}Madry et. al\\ PGD Train\end{tabular} & 99.4  & 0      \\ \hline
\multirow{5}{*}{\begin{tabular}[c]{@{}l@{}}With\\ Adv.\\ Training\end{tabular}}      & \multirow{3}{*}{\begin{tabular}[c]{@{}l@{}}Proposed\\ BP\end{tabular}}   & BN                                                               & 99.4  & 98.64  \\ \cline{3-5} 
                                                                                     &                                                                          & Tanh                                                             & 99.5  & 94.42  \\ \cline{3-5} 
                                                                                     &                                                                          & Tanh \& BN                                                       & 99.43 & 98.65  \\ \cline{2-5} 
                                                                                     & \multirow{2}{*}{\begin{tabular}[c]{@{}l@{}}Previous\\ Work\end{tabular}} & \begin{tabular}[c]{@{}l@{}}Thermometer\\ Encoding\end{tabular}   & 99.03 & 94.02  \\ \cline{3-5} 
                                                                                     &                                                                          & \begin{tabular}[c]{@{}l@{}}Madry et al.\\ PGD train\end{tabular} & 99.2  & 95.7   \\ \hline
\end{tabular}
\end{table}

Table 3 tabulates the accuracy in MNIST with different combination of blind pre-processing layers as discussed above.It gives more insight about their performance against the attack model. The experiment were performed using two different types of training. Clean training  is performed on original training data of MNIST. While adversarial training includes adversarial examples inside the training data. Both training result was evaluated using clean data and attack data. Clean data test basically tells how well the network performs the classification task when there is no attack. One of the key point here is to find which combination works as a better defense method, while causing clean data accuracy to go down marginally. However, by including adversarial training this loss of accuracy could be recovered. This is quite obvious as adversarial training increases the overall robustness of the network. Batch normalization turns out to be the most powerful tool in this defense when we do not include adversarial training. As a result tanh and smoothing filter fails without the support of batch normalization. The best result without adversary training could be obtained is 98.36 \% which is one of the major contribution of this defense. Previously, both thermometer encoding and Madry et al. found that, under PGD/LS-PGA attack, CNN accuracy drops to zero on MNIST data set without adverarial training.Moreover, the combinations that include BN \& Tanh and Smoothing \& BN also performed really well to defend against adversary without adversary training. Based on the experimental results, it could be concluded that adverarial training is not necessary using our proposed defense method for MNIST dataset, which reduces training overhead and keeps the accuracy on the clean data to 99.24 \%.

Additionally if adversarial training is included in our proposed method, the accuracy could further be improved. In this case, the combination of tanh function and batch normalization provides the best performance up to 98.65 \% result. Smoothing only helps to defend the attack when adversary training is not included. Next, we will conduct similar analysis in another two famous dataset, i.e. Cifar 10 and SVHN. 

\subsection{Performance in SVHN dataset}
The Street View House Numbers (SVHN) Dataset is a real world image for testing machine learning algorithms. For SVHN data set, we used Resnet-18 architecture \cite{he2016deep} in experiment and achieved state-of-the-art accuracy. Instead we only used tanh function for blind pre-processing on SVHN dataset. For this data set we used $\epsilon$ =0.047, $\delta=1.2$ xi=1 and step of attack = 10 as the the attack model parameter. We choose different attack model parameter for different data sets as reported by thermometer encoding defense. \cite{buckman2018thermometer} just to make the comparison fair.


\begin{table}[]
\centering
\caption{Result in SVHN dataset}
\label{my-label}
\begin{tabular}{|l|l|l|l|l|}
\hline
\begin{tabular}[c]{@{}l@{}}Train\\ Type\end{tabular}                                 & \begin{tabular}[c]{@{}l@{}}Work\\ Type\end{tabular}                    & Defense                                                 & Clean & Attack \\ \hline
\multicolumn{2}{|l|}{Original CNN}                                                                                                                            & No defense                                              & 95.1  & 6.99   \\ \hline
\multirow{4}{*}{\begin{tabular}[c]{@{}l@{}}Without \\ Adv. \\ Training\end{tabular}} & \multirow{3}{*}{\begin{tabular}[c]{@{}l@{}}Proposed\\ BP\end{tabular}} & BN                                                      & 81.16 & 47.71  \\ \cline{3-5} 
                                                                                     &                                                                        & Tanh                                                    & 96.56 & 49.39  \\ \cline{3-5} 
                                                                                     &                                                                        & Tanh \& BN                                              & 84.19 & 47.86  \\ \cline{2-5} 
                                                                                     & \begin{tabular}[c]{@{}l@{}}Previous\\ Work\end{tabular}                & \begin{tabular}[c]{@{}l@{}}Thermometer\\ /\end{tabular} & 96.96 & 48.02  \\ \hline
\multirow{4}{*}{\begin{tabular}[c]{@{}l@{}}With \\ Adv. \\ Training\end{tabular}}    & \multirow{3}{*}{\begin{tabular}[c]{@{}l@{}}Proposed\\ BP\end{tabular}} & BN                                                      & 45.50 & 93.48  \\ \cline{3-5} 
                                                                                     &                                                                        & Tanh                                                    & 96.11 & 95.06  \\ \cline{3-5} 
                                                                                     &                                                                        & Tanh \& BN                                              & 59.84 & 94.55  \\ \cline{2-5} 
                                                                                     & \begin{tabular}[c]{@{}l@{}}Previous\\ Work\end{tabular}                & \begin{tabular}[c]{@{}l@{}}Thermometer\\ /\end{tabular} & 97.18 & 95.02  \\ \hline
\end{tabular}
\end{table}

From table-4, we could easily observe the effect of this defense changes for SVHN dataset. Firstly, batch normalization still works as a strong defense but it causes the clean data accuracy to go down a lot. Tanh and Smoothing performs better than thermometer encoding without adversarial training. Looking at the clean data accuracy we propose to keep only the tanh function for defending against adversary against SVHN dataset. Our proposed defense method combined with adversarial training could achieve 95.06 \% accuracy. But adversarial training had to be used since we removed BN for SVHN data to encounter for the loss of accuracy on clean data. Batch normalization layer worked as the key component in the defense of MNIST without adversary training. However, after including the adversary training and blind pre-processing we could recover state of the art accuracy on SVHN data set as well.

So here one of the key problem of using batch normalization and smoohting is that they do sacrifice a lot of clean data accuracy. After using BN and smoohting in the pre-processing layer we see that clean data accuracy drops down to 45.50 \% and 69.12 \% respectively. As a result even though they provide some form of non-linearity but fail as a defense. Thus one of the key conclusion from this work is that if we find more non-linear function that do not sacrifice clean data accuracy then they might be a potential candidate as a robust defense.

\subsection{Performance in Cifar 10 dataset}

For Cifar 10 data set, we used Resnet-50 architecture with a little bit of modification on the attack strength. Even though thermometer encoding reported that using wider network helps in increasing the accuracy against adversarial examples and wide resnet is found to perform really well in defending adversary. However, in this paper we wanted to show the performance of functional filter in improving the accuracy from the previous defense model and we leave the analysis on the choice of architecture for future investigation. We choose the attack parameter for cifar 10 with $\epsilon$ equal to 0.031 and number of step of attack equal to 7.


\begin{table}[]
\centering
\caption{Result on CIFAR 10 dataset}
\label{my-label}
\begin{tabular}{|l|l|l|l|l|}
\hline
\begin{tabular}[c]{@{}l@{}}Train\\ Type\end{tabular}                                 & \begin{tabular}[c]{@{}l@{}}Work\\ Type\end{tabular}                    & Defense                                                        & Clean & Attack \\ \hline
\multicolumn{2}{|l|}{Original CNN}                                                                                                                            & No defense                                                     & 93.72 & 4.10   \\ \hline
\multirow{4}{*}{\begin{tabular}[c]{@{}l@{}}Without \\ Adv. \\ Training\end{tabular}} & \multirow{3}{*}{\begin{tabular}[c]{@{}l@{}}Proposed\\ BP\end{tabular}} & BN                                                             & 79.27 & 10.69  \\ \cline{3-5} 
                                                                                     &                                                                        & Tanh                                                           & 90.36 & 46.14  \\ \cline{3-5} 
                                                                                     &                                                                        & Tanh \& BN                                                     & 79.67 & 15.45  \\ \cline{2-5} 
                                                                                     & \begin{tabular}[c]{@{}l@{}}Previous\\ Work\end{tabular}                & \begin{tabular}[c]{@{}l@{}}Thermometer\\ Encoding\end{tabular} & 90.42 & 43.5   \\ \hline
\multirow{4}{*}{\begin{tabular}[c]{@{}l@{}}With \\ Adv. \\ Training\end{tabular}}    & \multirow{3}{*}{\begin{tabular}[c]{@{}l@{}}Proposed\\ BP\end{tabular}} & BN                                                             & 53.61 & 75     \\ \cline{3-5} 
                                                                                     &                                                                        & Tanh                                                           & 89.52 & 86.61  \\ \cline{3-5} 
                                                                                     &                                                                        & Tanh \& BN                                                     & 63.71 & 77.46  \\ \cline{2-5} 
                                                                                     & \begin{tabular}[c]{@{}l@{}}Previous\\ Work\end{tabular}                & \begin{tabular}[c]{@{}l@{}}Thermometer\\ Encoding\end{tabular} & 90.18 & 83.6   \\ \hline
\end{tabular}
\end{table}

Similar to SVHN we found good result using tanh as a pre-processing filter for cifar 10 data.  Additionally, it is observed that thermometer encoding produced a lower accuracy even without adversarial training. We could improve the accuracy without adversary training by 3\% while maintaining close to 90\% accuracy on clean data. After using adversarial training we could get close to 86.66\% accuracy on CIFAR 10. So regardless of data set, our proposed defense method helps in improving accuracy without adversary training. However, to achieve better accuracy, we need to include adversarial training. The accuracy reported in this paper is higher than the original thermometer encoding paper because we used a deeper architecture which makes the thermometer defense more robust.

After observing result for SVHN and CIFAR 10 it is clear that BN and smoothing causes a lot of sacrifice in clean data accuracy. This results in the model not learning properly during the training. As a result when we combine BN and Smoothing we get the worst result for this two datasets. THus we chose to keep only the tanh function in our proposed defense for CIFAR 10 and SVHN dataset. In the next section we analyze the expreimental result and try to explain their relative phenomenon.

\section{Analysis}

Based on the experimental results discussed in previous section, it shows that blind pre-processing can defend only MNIST data set without adversarial training. blind pre-processing can also defend against SVHN and CIFAR10 with adversarial training. We thus claim that the need for adversarial training for MNIST data set was eliminated. In this section we present some more in-depth analysis on MNIST data set to establish our claim. We also report black box attack accuracy in this section to prove the attacks strength even further.

In order to demonstrate the strength of our defense we show how far our defense can go in terms of defending in the advent of very strong attacks \cite{mkadry2017towards,buckman2018thermometer}. By changing the value of $\epsilon$ the strength of the attack can be modified. A larger value of $\epsilon$ indicates stronger attack. It was observed that the performance of previous defenses decreases exponentially with increasing attack strength as shown in table 5. However, BP defense strength decreases only a small amount with increasing attack strength is another proof of the success of BP as a universal robust defense.

\begin{table}[h]
\centering
\caption{Result of defense after increasing the attack strength-$\epsilon$ on MNIST}
\scalebox{0.8}{
\begin{tabular}{ |c|c|c|c| } 
 \hline
 Value of $\epsilon$ & BP WithoutAdv & BP WithAdv & Thermo WithAdv \\ 
 0.1 & 98.99 & 99.07 & 98.52 \\
 0.2 & 98.68 & 98.88 & 96.85 \\ 
 0.3 & 98.36 & 98.61 & 94.02 \\
 0.4 & 97.78 & 98.22 & 82.64 \\ 
 0.5 & 95.52 & 97.88 & 38.08 \\
 \hline
\end{tabular}}
\end{table}

As described earlier changing the value of $\epsilon$ would change the accuracy a lot since it decides how much the input image would be changed to cause the miss classifications. Our defense can withstand this power full attack even if the value of $\epsilon$ is increased to 0.5. No previous defense method, even under weaker attacks like FGSM, can withstand such large level of change against MNIST data set. What makes it more interesting is that this result can be achieved even without adversarial training. If this defense model combined with adversarial training then this would become the strongest defense ever reported against MNIST data set. On the other hand the accuracy thermometer encoding drops down to 38.08\% when $\epsilon$=0.5, even using adversarial training. We analyse this robustness of our defense against Carlini Wagner method now. The mean distance change to cause the miscalssification against carlini wagner attack is a metrice to evaluate e defenses robustness. 

\begin{table}[h]
\centering
\caption{Value of $\alpha$ on Different combination}
\begin{tabular}{ |c|c|c| } 
 \hline
 Defense & $\alpha$ & accuracy \\ 
 Tanh+BN & 49.8 & 98.65 \\ 
 Tanh+ Filter & 45.98 & 83.54 \\
 Filter+BN & 49.78 & 98.63 \\
 All three & 49.78 & 98.61 \\
 \hline
\end{tabular}
\end{table}

We also found that overall strength of the defense depends on the accuracy of clean data as well. In order to analyze it, we first define an important parameter called attacked data accuracy ratio \textit{($\alpha$)}. If we assume the accuracy on clean data is x\% and on attacked data is y\% then,
\begin{equation}
    \alpha= \frac{y}{(x+y)}*100\%
\end{equation}

We observed that defense combination that performed well in table 3 and 4 do have a higher value of $\alpha$. Lets look back at the same combination of table 3 for using adversarial training in table 7.

\begin{figure}[h]
\centering
\scalebox{1}{
\includegraphics[width=8cm]{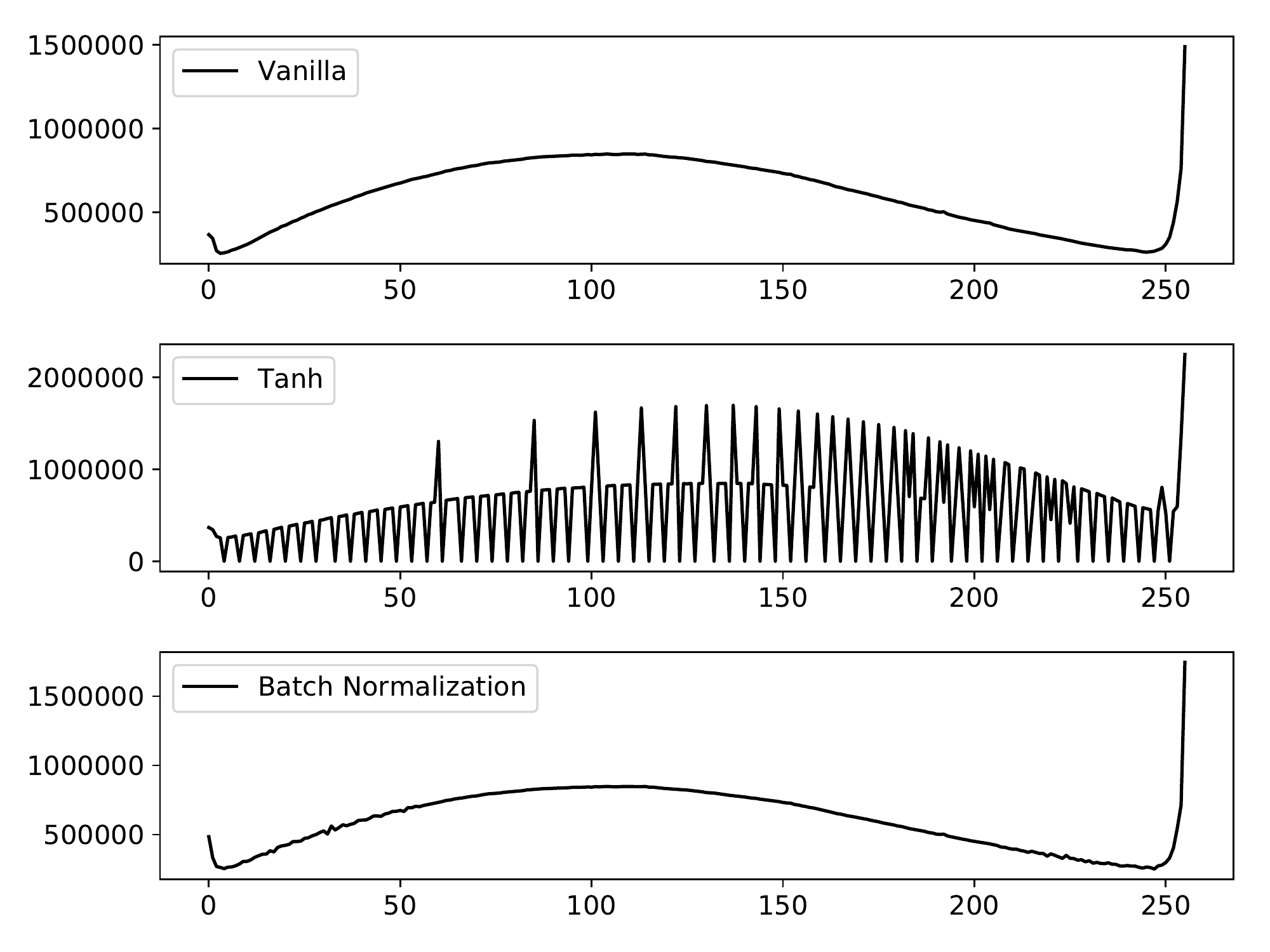}
}
\caption{CIFAR 10 dataset pixel frequency distribution by using different BP}
\end{figure}

The higher the accuracy, the higher the value of $\alpha$ in table 7 . It is also the same for svhn and cifar 10. The significance of increasing $\alpha$ is that, with increasing performance of the defense, the clean data accuracy actually goes down a little bit. Better defense performance usually comes with a little sacrifice of clean data accuracy. Since BN performed the best for MNIST it also maintained a greater value of $\alpha$ indicating it will hamper the clean data accuracy most. As MNIST is a small dataset the effect was negligible. However, for SVHN and CIFAR 10, this effect on accuracy degradation on clean data was severe. That's why we choose to drop BN from the model. As we found tanh optimizes between clean data accuracy and attacked data accuracy best for those two. We could also explain this phenomenon of BN not working against cifar by analysing the pixel intensity distribution of the dataset.


In order to understand the robustness of our defense more intuitively it is necessary to understand the frequency distribution of dataset pixels. As shown in fig. 4, the whole dataset's pixel distribution completely changes after applying blind pre-processing. We understand that this redistribution in the image pixels plays an important role in the robustness of the defense. In fig. 4 typical CIFAR 10 dataset frequency distribution is displayed with pixel values on the x axis and their appearing frequency in the y axis. This distribution's standard deviation changes completely after applying Tanh processing. The key point is before applying blind pre-processing the pixel values are clustered around a certain pixel range. After processing with tanh their pixel standard deviation increases and becomes more scattered. As a result the pixel difference between the edges of the digit or object becomes larger. Thus fooling the network becomes harder as introducing noises will have less impact on the images due to larger transition near the edges. However, for CIFAR 10 BN does not change the distribution much like MNIST. The inability to change this distribution is one of the reason for the failure of BN in CIFAR 10 dataset.

\section{Summary}
We presented a robust defense method that uses a combination of pre-processing layers to defend against adversarial attacks. We choose tanh and batch normalization as our blind pre-processing layer. This processing layer is completely unknown and not accessible to the attacker. We could defend MNIST data set for the first time without using adversarial training. That reduces training time and complexity in the defense model. A combination of tanh and batch normalization worked in improving MNIST accuracy to close to 98\% from zero without adversarial examples. Moreoever, we showed that increasing attack defense where other defenses fail, our defense remains one of the strongest ever reported. Additionally the proposed BP model worked better than recent defenses against SVHN and CIFAR 10 as well. But, it required adversarial training to recover the accuracy upto 95\% for SVHN and 86\% for Cifar 10. We also showed under varying attack strength our defense would outperform its recent counter part. 

\section{Discussion}
Currently, we are conducting experiments on a variety of attack. We got more than good result in defending FGSM, Carlini Wagner attack and also on black box box. We are also the first to defend against FGSM attack without adversarial training reaching close to 96 \% accuracy. Again for Carlini Wagner attack our defense model requires highest amount input perturbation to miss-classify the input for all three distance metrices. One of the strongest attack recently released succeeded in breaking all the ICLR 2018 defend models. Even though we tested against obfuscated gradient attack \cite{obfuscated-gradients} and got reasonable amount of success but our concept of blind pre-processing actually makes their attack void.  
\section{Acknowledgements}
We would like to thank Jifeng Yi for insightful discussions.


\bibliography{reference}
\bibliographystyle{icml2018}

\end{document}